\documentclass[11pt,a4paper]{article}
\usepackage[hyperref]{emnlp2020}
\usepackage{times}
\usepackage{latexsym}

\usepackage{microtype}

\aclfinalcopy 

\usepackage{amsmath}
\usepackage{amsfonts}
\usepackage{amssymb}
\usepackage{amsthm}
\usepackage{graphicx}

\usepackage{multirow}
\title{Probabilistic Predictions of People Perusing: Evaluating Metrics of Language Model Performance for Psycholinguistic Modeling}

\author{Yiding Hao, Simon Mendelsohn, Rachel Sterneck,\\\textbf{Randi Martinez,} \and \textbf{Robert Frank} \\
  Yale University, New Haven, CT, USA \\
  \texttt{firstname.lastname@yale.edu} \\}

\date{}

\begin{document}
    \maketitle
    \begin{abstract}
        By positing a relationship between naturalistic reading times and information-theoretic surprisal, surprisal theory  \citep{haleProbabilisticEarleyParser2001,levyExpectationbasedSyntacticComprehension2008} provides a natural interface between language models and psycholinguistic models. This paper re-evaluates a claim due to \citet{goodkindPredictivePowerWord2018} that a language model's ability to model reading times is a linear function of its perplexity. By extending \citeauthor{goodkindPredictivePowerWord2018}'s analysis to modern neural architectures, we show that the proposed relation does not always hold for Long Short-Term Memory networks, Transformers, and pre-trained models. We introduce an alternate measure of language modeling performance called \textit{predictability norm correlation} based on Cloze probabilities measured from human subjects. Our new metric yields a more robust relationship between language model quality and psycholinguistic modeling performance that allows for comparison between models with different training configurations.
    \end{abstract}
    
    \section{Introduction}
    
    Naturalistic reading times are known to be affected by incongruities between the current word and the context created by preceding words (\citealp{zolaEffectRedundancyPerception1981,zolaRedundancyWordPerception1984,ehrlichContextualEffectsWord1981}, \textit{inter alia}). Words that are unexpected within their contexts are fixated for longer periods, while predictable words are fixated for a shorter amount of time or skipped altogether. This observation has been described formally by the suprisal theory of sentence processing \citep{haleProbabilisticEarleyParser2001,levyExpectationbasedSyntacticComprehension2008}, which posits that the difficulty of processing a particular word is directly proportional to its \textit{surprisal}, defined as the negative logarithm of its probability given its preceding context. 
    
    By relating psychometric observations with information-theoretic concepts, surprisal theory provides a natural bridge between psycholinguistic modeling on the one hand and language modeling on the other. While modeling studies for reading times traditionally use surprisals obtained from probabilistic parsers \citep{haleProbabilisticEarleyParser2001,dembergDataEyetrackingCorpora2008,roarkDerivingLexicalSyntactic2009} or directly from probabilistic context-free grammars \citep{levyExpectationbasedSyntacticComprehension2008,bostonParsingCostsPredictors2008}, surprisals from $n$-gram models \citep{mitchellSyntacticSemanticFactors2010,smithEffectWordPredictability2013a} and simple recurrent network language models \citep{frankSurprisalbasedComparisonSymbolic2009,monsalveLexicalSurprisalGeneral2012} have also been incorporated into reading-time studies. More recent work has sought to leverage the advances in language modeling made possible by neural NLP in order to determine whether modern techniques can yield more reliable estimates of surprisal for cognitive modeling. These studies have investigated the psycholinguistic capabilities of several neural architectures, including Long Short-Term Memory networks and Gated Recurrent Unit networks \citep{aurnhammerComparingGatedSimple2019}, Transformers \citep{merkxComparingTransformersRNNs2020}, and GPT-2 \citep{wilcoxPredictivePowerNeural2020}. In general, language models achieving a better \textit{perplexity} have been found to yield better psycholinguistic models via the surprisal estimates they furnish, with  \citet{goodkindPredictivePowerWord2018} proposing a linear relationship between the two factors.
    
    This paper revisits \citeauthor{goodkindPredictivePowerWord2018}'s proposed linear relationship between perplexity and psycholinguistic modeling performance. We address two drawbacks of their analysis, which hinder its applicability to the state of the art in language modeling. Firstly, \citeauthor{goodkindPredictivePowerWord2018} only consider $n$-gram models and a small LSTM language model with 50 hidden units. These constitute a small and relatively weak collection of language models compared to the diverse array of techniques and vast computing resources that are available today. Secondly, the reliability of perplexity as a measure of language modeling performance is highly dependent on training configurations, such that only perplexities for models with the same vocabulary are comparable (see \citealp[pp. 95--97]{jurafskySpeechLanguageProcessing2008}). This is especially problematic given the rise of large-scale, general-purpose \textit{pre-trained} language models, since vocabulary cannot be controlled for when comparing different pre-trained models.
    
    To address these issues, we first expand \citeauthor{goodkindPredictivePowerWord2018}'s empirical coverage by extending their analysis to Long Short-Term Memory networks (LSTMs, \citealp{hochreiterLongShortTermMemory1997}) and Transformers \citep{vaswaniAttentionAllYou2017} of varying sizes and training corpora, along with four large pre-trained models. We then propose an alternate metric of language modeling performance based on word probabilities obtained from humans using a Cloze procedure \citep{taylorClozeProcedureNew1953}, which allows different pre-trained models to be compared with one another despite differences in vocabulary. In line with previous work, we find that the relationship between perplexity and psycholinguistic modeling performance is weaker for LSTMs and Transformers than for $n$-gram models, and not consistent across vocabularies. However, our Cloze-based metric yields a relationship which is robust to differences in model vocabulary and incorporates both our trained and our pre-trained models. Based on this, we argue that our Cloze-based metric is more suitable than perplexity for evaluating pre-trained models, and better reveals the relationship between language modeling performance and psycholinguistic modeling performance.
    
    The structure of this paper is as follows. \autoref{sec:psychmodels} describes our procedure for modeling reading times, and \autoref{sec:langmodels} describes our language models as well as the Cloze-based performance metric. Our results are presented in \autoref{sec:results} and discussed in \autoref{sec:discussion}. \autoref{sec:conclusion} concludes.
    
    
    \section{Psycholinguistic Modeling}
    \label{sec:psychmodels}
    
    We follow the methodology of \citet{goodkindPredictivePowerWord2018} for psycholinguistic modeling. We model reading times using generalized additive mixed-effect models (GAMMs), as implemented in the \texttt{mgcv} R package \citep{woodStableEfficientMultiple2004}. Our GAMMs take several variables as input, optionally including language model surprisals. We fit a GAMM for each language model, along with a baseline model that does not include surprisal values. We measure the psycholinguistic modeling performance of each model by a score called \textit{delta log likelihood} ($\Delta$LogLik), defined as the difference in log likelihood between the model's corresponding GAMM and the baseline GAMM.
    
    In the following subsections, we describe the reading time data used in our study and our procedure for fitting GAMMs.
    
    \subsection{Eyetracking Data}
    
    The data for our psycholinguistic models come from the English portion of the Dundee Corpus \citep{kennedyDundeeCorpus2003a,kennedyDundeeCorpus2003}, a dataset containing naturalistic reading times from an eyetracking study. The data were elicited from ten English-speaking participants reading editorials from British newspaper \textit{The Independent}. The editorials were divided into several \textit{texts}, which were shown to subjects on a screen one at a time. Reading times were obtained for a total of 56,212 tokens, drawn from a vocabulary of 9,776 unique word types, excluding punctuation. Punctuation marks were not treated as separate tokens, but rather as belonging to the words they are attached to.
    
    \subsection{GAMMs}
    \label{sec:gamm}
    
    Following \citet{goodkindPredictivePowerWord2018} and \citet{smithEffectWordPredictability2013a}, our GAMMs consist of the following input terms:
    \begin{itemize}
        \item linear terms for the surprisal of the current and previous word;
        \item tensor product interactions\footnote{This is a 2-dimentional tensor spline interaction.} between word length and log (unigram) frequency for the current and previous word;
        \item a spline\footnote{Penalized spline regression uses a high-dimensional spline basis to estimate unknown non-linear relations. In order to avoid the over-fitting that otherwise plagues such high-dimensional models, it combines
the standard maximum likelihood criterion with a curvature penalty term that biases the regression towards smoother curves.} term for the current word's position within a text; and
        \item a binary term representing whether or not the previous word was fixated.
    \end{itemize}
    The linear terms for surprisal are excluded from the baseline GAMM. Unigram frequencies were estimated using the One Billion Word Benchmark (LM1B, \citealp{chelbaOneBillionWord2014}). We use the same preprocessing procedure as \citeauthor{goodkindPredictivePowerWord2018}, which removes from the data the first and last words of each text, words preceding punctuation marks, words containing non-alphabetic characters, words for which no unigram frequency estimate is available, and words following these words. In cases where a word from the Dundee Corpus is tokenized into multiple tokens by a language model, we take the word's surprisal to be the sum of the surprisals for each of the constituent tokens.
    
    \citeauthor{goodkindPredictivePowerWord2018}'s GAMM outputs represent predictions for \textit{gaze duration} (GD), defined as the time elapsed between the first fixation on a word token and the first time the subject exits that token. In addition to gaze duration, we train GAMMs to predict two other eyetracking measures: \textit{first fixation duration} (FFD), defined as the time elapsed during the first fixation on a token, and \textit{total duration} (TD), defined as the total amount of time spent looking at a token.
    
    \section{Language Modeling}
    \label{sec:langmodels}
    
    Traditionally, language models form independent components of NLP systems, their outputs serving as inputs to downstream applications. However, following recent advances in transfer learning for NLP, modern neural language models are often designed not to compute word probabilities \textit{per se}, but rather to provide representations of linguistic knowledge that can facilitate training on other tasks \citep{daiSemisupervisedSequenceLearning2015,howardUniversalLanguageModel2018}. Under this paradigm, a single large neural network is \textit{pre-trained} on a language modeling or other word prediction objective. This pre-trained language model can then be trained, or \textit{fine-tuned}, on another task such as text classification \citep{yangXLNetGeneralizedAutoregressive2019} or machine translation \citep{conneauCrosslingualLanguageModel2019}. Both kinds of language models are considered in this paper.
    
    \subsection{Traditional Language Models}
    
    \begin{table}
        \small
        \centering
        \begin{tabular}{l c c }
            \hline
            \textbf{Model} & \textbf{Penn Treebank} & \textbf{WikiText-2} \\\hline
            2-Gram & 185.7 & 248.5 \\
            3-Gram & 148.3 & 211.1 \\
            4-Gram & 142.7 & 206.0 \\
            5-Gram & \textbf{141.2} & \textbf{204.8} \\\hline
            LSTM-200 & 89.3 & 106.0 \\
            LSTM-425 & \textbf{85.1} & \textbf{98.5} \\
            LSTM-650 & 89.8 & 98.8 \\
            LSTM-1100 & 97.9 & 110.5 \\\hline
            Transformer-200 & \textbf{117.8} & \textbf{150.2} \\
            Transformer-424 & 119.1 & 151.5 \\
            Transformer-650 & 118.7 & 154.7 \\
            Transformer-1100 & 119.7 & 156.6 \\\hline
        \end{tabular}
        \caption{Perplexities attained by our trained models on the Penn Treebank and WikiText-2 testing sets.}
        \label{tab:testing_ppl}
    \end{table}
    
    We consider three types of traditional language models: $n$-gram models, LSTM language models, and Transformer language models. We train four models of each type, with varying numbers of parameters, on the Penn Treebank corpus \citep{mikolovEmpiricalEvaluationCombination2011} and the WikiText-2 corpus \citep{merityPointerSentinelMixture2016}. We additionally train $n$-gram models on LM1B in order to reproduce and facilitate comparison with \citeauthor{goodkindPredictivePowerWord2018}'s (\citeyear{goodkindPredictivePowerWord2018}) results.
    
    Following \citet{goodkindPredictivePowerWord2018}, our $n$-gram models were trained using modified Kneser--Ney smoothing with KenLM \citep{heafieldKenLMFasterSmaller2011,heafieldScalableModifiedKneserNey2013}. We trained $n$-gram models with $n = \text{2}$, 3, 4, and 5 on each dataset. We additionally trained a unigram model in order to populate the word frequency term in the input to the GAMMs. 
    
    Our LSTM and Transformer models are based on an implementation publicly available on the official PyTorch website \citep{paszkeAutomaticDifferentiationPyTorch2017}.\footnote{\url{https://github.com/pytorch/examples/tree/master/word_language_model}} Both models include an embedding layer and a softmax decoding layer. The LSTM model consists of two LSTM layers, while the Transformer model consists of two encoder blocks with masked self-attention, each containing two attention heads. We trained LSTM models with 200, 425, 650, and 1100 hidden units (LSTM-200, -425, -650, and -1100, respectively), as well as Transformer models with 200, 424,\footnote{Because our Transformer model has two attention heads, it needs to have an even number of hidden units, so we used 424 instead of 425 (like we used in the LSTM).} 650, and 1100 hidden units (Transformer-200, -424, \mbox{-650}, and -1100, respectively). The embedding size used by each model was equal to its hidden size. For each architecture, training corpus, and hidden size, we trained a model using SGD with the learning rate annealed by .5 when perplexity does not improve on a validation set. Training occurred for a maximum of 50 epochs, stopping early with a patience of 5. The batch size, dropout rate, and initial learning rate were tuned using a Bayesian optimization routine consisting of 7 random trials followed by 13 GPEI trials \citep{snoekPracticalBayesianOptimization2012}. The LSTM-1100 and Transformer-1100 models were trained using the best hyperparameters for LSTM-650 and Transformer-650, respectively. The testing perplexities attained by our trained models are shown in \autoref{tab:testing_ppl}.
    
    \subsection{Pre-Trained Models}
    
    In addition to our traditional language models, we include four pre-trained models in our analysis: GPT-2 \citep{radfordLanguageModelsAre2019}, XLM \citep{conneauCrosslingualLanguageModel2019}, Transformer-XL \citep{daiTransformerXLAttentiveLanguage2019},  and XLNet \citep{yangXLNetGeneralizedAutoregressive2019}. All four models are variants of Transformer language models but differ from one another in terms of training setup and architectural details. We use the implementations from Hugging Face's \textit{Transformers} library \citep{wolfHuggingFaceTransformersStateoftheart2020} off the shelf, with no fine-tuning. We briefly describe the distinguishing features of each model below.
    
    \subsubsection{GPT-2}
    
    GPT-2 is a Transformer language model trained on a large corpus called WebText, which was designed to include a diverse array of documents in order to capture domain-general linguistic knowledge. The dataset consists of roughly 8 million webpages obtained from Reddit links. It demonstrates state of the art perplexities on various language modeling testing sets, as well as a notable ability to generate human-like text, especially in the context of summarization. \citet{radfordLanguageModelsAre2019} present GPT-2 models in four sizes; we use the smallest size, which consists of 12 layers, 12 attention heads, and 768 hidden units. GPT-2's vocabulary uses a byte-pair encoding, with 50,257 unique word and sub-word types.
    
    \subsubsection{XLM}
    
    XLM is similar in approach to GPT-2, but it is specifically designed for cross-lingual language tasks, including multilingual classification and machine translation. The XLM model we use was trained on a combination of English and German Wikipedia entries. It has 6 layers, 1024 hidden units, and 8 attention heads. XLM also uses a byte-pair encoding in its vocabulary, with 64,699 unique word and subword types. Unlike the other models, however, XLM's vocabulary includes German tokens in addition to English tokens.
    
    \subsubsection{Transformer-XL}
    
    The Transformer-XL model introduces architectural enhancements to the Transformer that facilitate learning of long-distance dependencies. It does this using relative encodings and a segment recurrence mechanism, which augments the Transformer with elements of RNN language models. Our Transformer-XL model was trained on the WikiText-103 corpus \citep{merityPointerSentinelMixture2016}, and contains 18 layers, 1024 hidden units, and 16 attention heads.
    
    \subsubsection{XLNet} 
    
    Finally, XLNet is a language model that extends Transformer-XL to account for bidirectional context within a language modeling setting. This is achieved using a generalized autoregressive technique, in which words are predicted in a randomized order for each training batch. We use an XLNet model with 12 layers, 768 hidden units, and 12 attention heads, trained on a custom corpus drawn from various sources. 

    \subsection{Language Model Evaluation}
    
    We employ two methods for evaluating the outputs of our language models: \textit{perplexity}, the standard evaluation metric for language modeling, and \textit{predictability norm correlation}, our proposed metric for comparing models with different vocabularies. 
    
    \subsubsection{Perplexity}
    
    For a given language model, the perplexity of a text consisting of $N$ tokens is defined by the formula
    \[ 
    \text{perplexity} = \left(\prod_{i = 1}^N P(\text{token}_i|\text{tokens}_{j<i}) \right)^{-\frac{1}{N}}
    \]
    where $P(\text{token}_i|\text{tokens}_{j<i})$ is the probability assigned to the $i$th token after the model has processed the first $i - 1$ tokens. Perplexity can also be defined as the exponential of the average surprisal of the text.
    \[
    \text{perplexity} = e^{-\frac{1}{N}\sum_{i = 1}^N \ln(P(\text{token}_i|\text{tokens}_{j<i}))}
    \]
    Intuitively, perplexity may be interpreted as the weighted average number of possibilities the language model chooses between when predicting the words in the text. Lower perplexities indicate better language modeling performance, since the model is less uncertain about its predictions. Note that using a larger vocabulary artificially increases perplexity, since the model automatically has more words to choose from, thus decreasing $P(\text{token}_i|\text{tokens}_{j<i})$ on average.
    
    In this study, we calculate the perplexity of each language model on the entire Dundee Corpus, without the preprocessing of \autoref{sec:gamm}. All perplexity calculations are based on the tokenization used in \citet{goodkindPredictivePowerWord2018}, which divides the Dundee Corpus into $N = \text{60,916}$ tokens. This produces slightly more tokens than the original tokenization used by \citet{kennedyDundeeCorpus2003a}, since punctuation marks are treated as separate tokens, following common practice in language modeling.
    
    \subsubsection{Predictability Norm Correlation}
    
    As an alternative to perplexity, we evaluate models according to a \textit{predictability norm correlation score} (PNC), defined as the Pearson correlation between surprisal values computed by a language model and surprisal values measured from human subjects using a Cloze task. The data used to calculate PNC come from predictability norms collected by \citet{kennedyFrequencyPredictabilityEffects2013} for a 16-sentence subset of the Dundee Corpus. Each subject was shown a random and possibly empty initial segment of each sentence and asked to predict the next word, typing their response into a computer. Predictions were collected from 272 subjects in total, with roughly 25 predictions for each token. The ``human'' probability values are defined by the proportion of responses for each token that represent correct predictions.  \citet{kennedyFrequencyPredictabilityEffects2013} report two sets of human probability values: one that counts minor misspellings of the target word as correct predictions, and one that counts them as incorrect. We use the former set of scores.

    \begin{table}[htb]
        \small
        \centering
        \begin{tabular}{l c c | c c c}
            \hline
            \multirow{2}{*}{\textbf{Model}} & \multirow{2}{*}{\textbf{PPL}} & \multirow{2}{*}{\textbf{PNC}} & \multicolumn{3}{c}{$\Delta$\textbf{LogLik}} \\
            & & & \textbf{FFD} & \textbf{GD} & \textbf{TD} \\\hline
            
            \multicolumn{3}{l|}{\textsc{Pre-Trained}} \\
            GPT-2 & \textbf{87.6} & \textbf{.633} & \textbf{180.9} & \textbf{332.7} & \textbf{841.2} \\
            XLM & 410.3 & .155 & 35.7 & 95.6 & 131.3 \\
            Trans.-XL & 152.6 & .566 & 103.9 & 183.7 & 493.7 \\
            XLNet & 489.2 & .580 & 141.5 & 259.7 & 646.0 \\\hline
            
            \multicolumn{3}{l|}{\textsc{Penn Treebank}}  \\
            2-Gram & 216.9 & .300 & 22.4 & 27.6 & 69.0 \\
            3-Gram & 207.0 & .326 & 24.1 & 33.2 & 95.1 \\
            4-Gram & 206.1 & .330 & 22.5 & 32.4 & 95.6 \\
            5-Gram & 205.7 & \textbf{.331} & 22.4 & 32.3 & 96.3 \\
            LSTM-200 & 121.4 & .270 & 49.7 & 93.9 & 226.8 \\
            LSTM-425 & \textbf{118.4} & .271 & 51.3 & 94.4 & 238.1 \\
            LSTM-650 & 125.5 & .273 & \textbf{60.8} & \textbf{107.7} & \textbf{250.3} \\
            LSTM-1100 & 123.3 & .270 & 47.1 & 87.9 & 222.3 \\
            Trans.-200 & 137.5 & .277 & 49.2 & 88.0 & 215.6 \\
            Trans.-424 & 144.1 & .274 & 47.8 & 84.5 & 209.0 \\
            Trans.-650 & 141.4 & .275 & 51.4 & 89.0 & 221.8 \\
            Trans.-1100 & 150.7 & .278 & 44.1 & 81.8 & 211.4 \\\hline
            
            \multicolumn{3}{l|}{\textsc{WikiText-2}} \\
            2-Gram & 381.1 & .425 & 19.3 & 32.6 & 97.0 \\
            3-Gram & 364.4 & .453 & 28.2 & 45.0 & 125.4 \\
            4-Gram & 359.8 & .455 & 28.2 & 45.5 & 126.6 \\
            5-Gram & 358.6 & .455 & 28.2 & 45.9 & 127.3 \\
            LSTM-200 & 236.5 & .443 & 52.1 & 93.0 & 255.7 \\
            LSTM-425 & \textbf{224.4} & .437 & 49.2 & 84.8 & 243.4 \\
            LSTM-650 & 230.5 & .436 & 50.6 & 84.9 & 240.6 \\
            LSTM-1100 & 262.3 & .450 & 47.2 & 84.0 & 250.9 \\
            Trans.-200 & 319.1 & .443 & 54.4 & 92.9 & 251.3 \\
            Trans.-424 & 329.4 & \textbf{.462} & \textbf{61.6} & \textbf{104.5} & \textbf{279.5} \\
            Trans.-650 & 340.7 & .450 & 53.6 & 95.5 & 271.9 \\
            Trans.-1100 & 337.1 & .454 & 53.9 & 92.6 & 261.2\\\hline
            
            \multicolumn{3}{l|}{\textsc{LM1B}}  \\
            2-Gram & 291.1 & .506 & 86.9 & 149.1 & 413.6 \\
            3-Gram & 191.2 & .560 & 122.1 & 212.2 & 546.8 \\
            4-Gram & 172.2 & .582 & 130.5 & 220.4 & 552.1 \\
            5-Gram & \textbf{169.0} & \textbf{.583} & \textbf{131.3} & \textbf{223.4} & \textbf{553.9} \\\hline
        \end{tabular}
        \caption{Perplexity (PPL), PNC, and $\Delta$LogLik.}
        \label{tab:results}
    \end{table}
    
    \section{Results}
    \label{sec:results}
    
    \begin{figure*}
        \centering
        \includegraphics{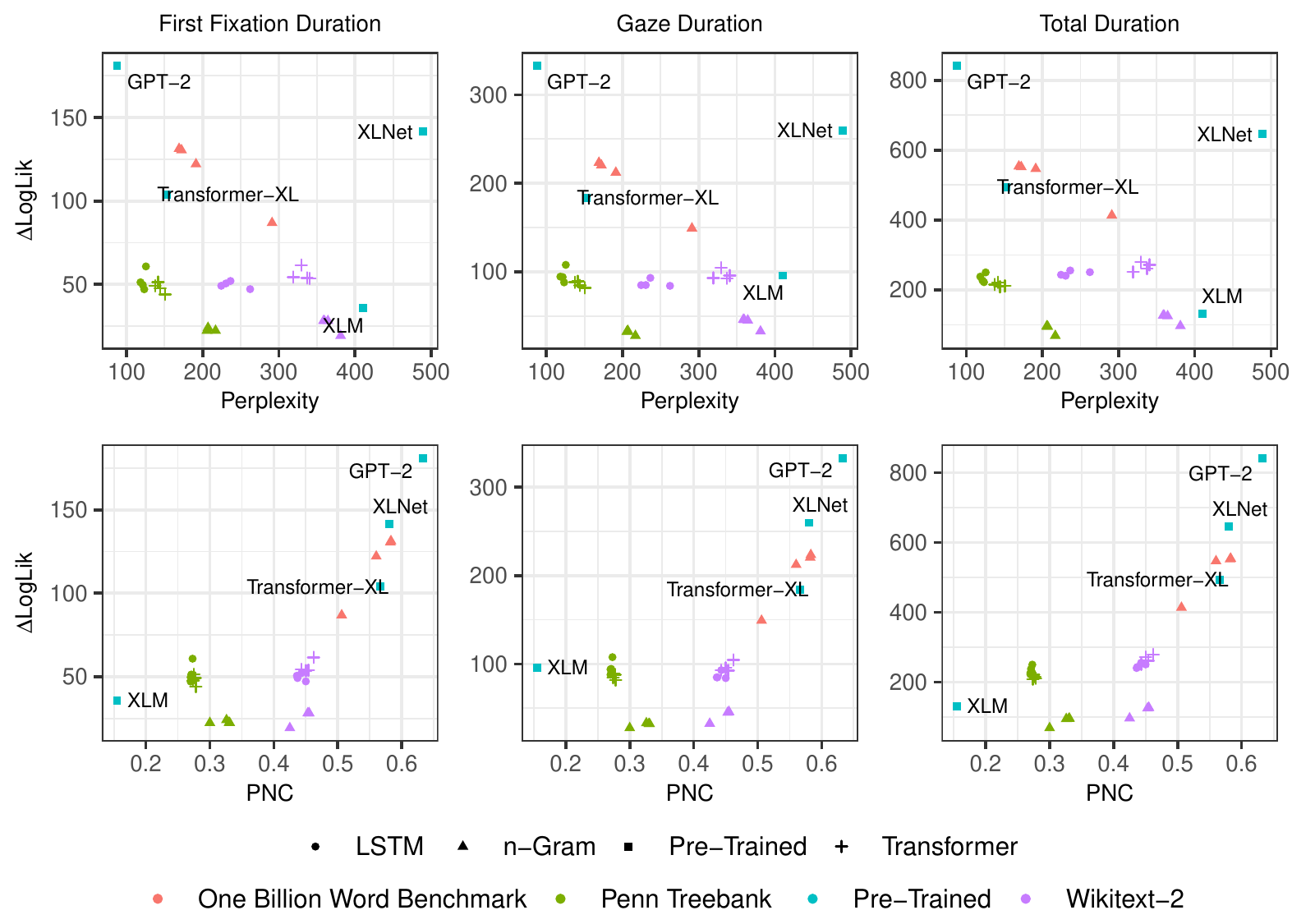}
        \caption{The relationship between language modeling performance and $\Delta$LogLik.}
        \label{fig:results}
    \end{figure*}
    
    Our results are presented in \autoref{tab:results} and \autoref{fig:results}, which show the same data in tabular and graphical form. Overall, GPT-2 outperforms all other models on all metrics, achieving the best perplexity, PNC, and $\Delta$LogLik. Transformer-XL, XLNet, and the $n$-gram models trained on LM1B perform significantly better in terms of $\Delta$LogLik than XLM and the models trained on Penn Treebank and WikiText-2. This may be because Penn Treebank and WikiText-2 are significantly smaller than the other training datasets, and because XLM, being a multilingual model, is not as suited to modeling data from native English speakers as the other models.
    
    The models trained on Penn Treebank and WikiText-2 achieve similar levels of $\Delta$LogLik, though the Penn Treebank models generally have a better perplexity while the WikiText-2 models generally have a better PNC. Between LSTMs and Transformers, neither architecture is consistently better than the other in terms of $\Delta$LogLik: Penn Treebank LSTMs outperform Penn Treebank Transformers on average, while WikiText-2 Transformers outperform WikiText-2 LSTMs on average. However, controlling for training corpus, our LSTMs are consistently better than our Transformers in terms of perplexity, while our Transformers consistently outperform our LSTMs in terms of PNC.
    
    In the remainder of this section, we describe specific observations about the relationship between language model performance and $\Delta$LogLik. 
    
    \subsection{Perplexity vs. $\Delta$LogLik}
    \label{sec:ppl_deltaloglik}
    
    The relationship between perplexity and $\Delta$LogLik is visualized in the top row of \autoref{fig:results}. Recall that only models trained on the same corpus can be compared with one another in these plots, since the different training corpora have different vocabularies. Indeed, we see a large effect overall of training data on the relationship between perplexity and $\Delta$LogLik. Apart from GPT-2 and Transformer-XL, observe that the models trained on the Penn Treebank achieve the lowest perplexities. This likely reflects the fact that the Penn Treebank has the smallest vocabulary among the different models, as well as the fact that the Penn Treebank and the Dundee Corpus are both drawn from newspapers.
    
    For all three eyetracking measures, there appears to be a linear relationship between perplexity and $\Delta$LogLik for different sized $n$-gram models with the same training corpus, as well as the models trained on the Penn Treebank. However, this relationship does not generalize well to the LSTMs and Transformers trained on WikiText-2. Neither the LSTMs nor the Transformers extrapolate the line that connects the $n$-gram models. In agreement with \citet{goodkindPredictivePowerWord2018}, we observe that the LSTMs lie below the $n$-gram line, while the Transformers lie above it. Among models of the same training corpus and architecture, we do not see any relationship between perplexity and $\Delta$LogLik for LSTMs or Transformers trained on WikiText-2.
    
    Because they all use different vocabularies, the pre-trained models cannot be compared with other models in these plots. Nonetheless, XLNet appears to be an outlier among pre-trained models, having the highest perplexity despite achieving the second-highest $\Delta$LogLik for all three eyetracking measures.
    
    Looking across the columns of \autoref{fig:results}, we observe that the results described in this subsection generalize across all three reading time metrics for the Dundee Corpus.  
    
    \subsection{PNC vs. $\Delta$LogLik}
    \label{sec:pnc_deltaloglik}

    Next, let us turn to the relationship between PNC and $\Delta$LogLik, visualized on the bottom row of \autoref{fig:results}. We see a robust relationship in the data, especially among the better models. In particular, the models that achieve a PNC of at least 0.4 show a strong linear relationship between PNC and $\Delta$LogLik. Alternatively, the results may be seen as an exponential or logistic relation that subsumes the models with $\text{PNC} < \text{0.4}$, as $\Delta$LogLik cannot be negative and PNC is capped at 1. Crucially, all of the models with $\text{PNC} \geq \text{0.4}$ are subsumed under the \textit{same trend}, despite the fact that these models collectively use five different vocabularies of differing sizes. Furthermore, unlike in the case of perplexity, LSTMs do not overperform in terms of PNC relative to their $\Delta$LogLik performance.
    
    Among the pre-trained models, the outlying data point is the XLM model, which achieves the lowest PNC out of all the models. This is unsurprising, considering that XLM is a multilingual model. The other potential outliers are the Penn Treebank data points, which achieve PNCs below 0.4 despite having a similar $\Delta$LogLik to the WikiText-2 models.  
    
    As with perplexity, these results generalize across the reading metrics for the Dundee Corpus, though visually speaking, $\Delta$LogLiks based on first fixation duration and total duration appear to provide a bitter fit for an exponential model than gaze duration.
    
    \section{Discussion}
    \label{sec:discussion}
    
    \begin{figure}
        \centering
        \includegraphics[scale=.29]{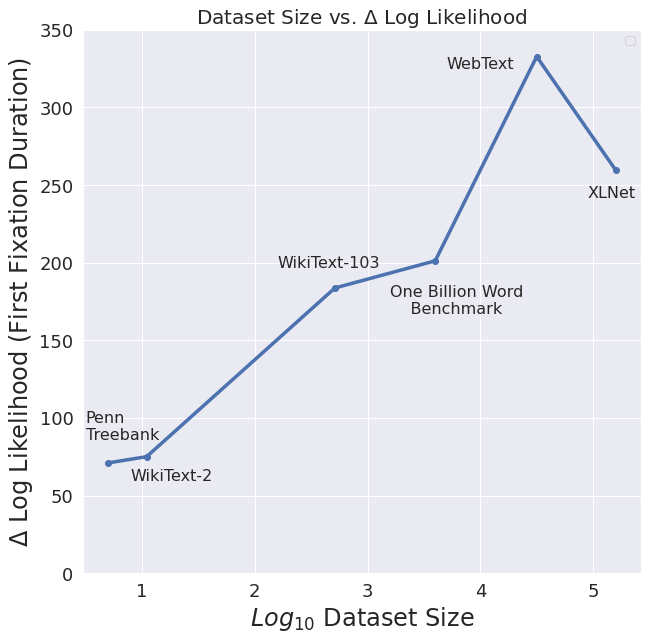}
        \caption{The relationship between training dataset size (MB) and $\Delta$LogLik for monolingual English models. Values are averaged among models trained on the each dataset.}
        \label{fig:datasetsize_deltaloglik}
    \end{figure}

    \begin{figure}
        \centering
        \includegraphics[scale=.29]{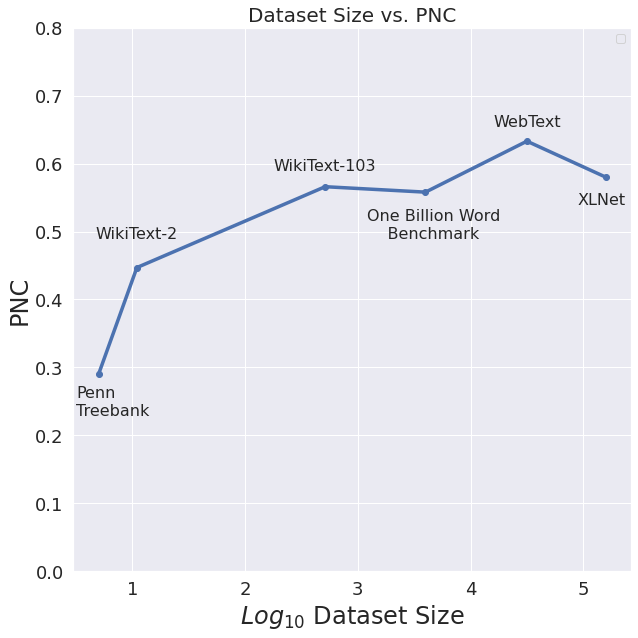}
        \caption{The relationship between training dataset size (MB) and PNC for monolingual English models. Values are averaged among models trained on the each dataset.}
        \label{fig:datasetsize_pnc}
    \end{figure}
    
    We focus our discussion on three topics. First, we consider the ways in which each of our experimental variables affects our models' psycholinguistic modeling performance. Next, we briefly identify some properties of PNC as a metric, particularly in terms of its relationship with training corpus size. Finally, we reflect on the implications of our results for language modeling and psycholinguistic modeling evaluation more generally.
    
    \subsection{Factors Affecting $\Delta$LogLik}
    In our experiment, we have analyzed several model architectures, model sizes, and training datasets. Similarly to \citet{merkxComparingTransformersRNNs2020}, we determine that the number of model parameters generally does not significantly impact psycholinguistic modeling performance, whereas model architecture, along with the composition and size of the training corpus, do have a significant impact. 
    
    Our findings regarding the effect of model architecture are consistent with the model class effect identified by both \citet{goodkindPredictivePowerWord2018} and \citet{wilcoxPredictivePowerNeural2020}, in that LSTMs appear to underperform in terms of $\Delta$LogLik given their perplexities. However, our architecture effect is not as dramatic as the one reported by \citeauthor{wilcoxPredictivePowerNeural2020}: whereas their $n$-grams generally show superior psychometric predictive power over their LSTMs, our $n$-gram models are not on par with that of our LSTMs. 
    
   Although our experiment did not control for both training data size and composition separately, we argue that both properties are important factors that affect $\Delta$LogLik. Firstly, notice that the $\Delta$LogLiks for the Penn Treebank LSTMs and Transformers are exceptionally high given that their PNC is less than 0.4, especially the values computed for total duration. Indeed, whereas all other models with $\text{PNC} < \text{0.4}$ achieve $\Delta$LogLiks close to 100 for total duration, the Penn Treebank LSTMs and Transformers exhibit $\Delta$LogLiks in excess of 200. We posit that this phenomenon is due to domain similarity between the newspaper data found in the Penn Treebank and Dundee Corpus datasets, suggesting that the content of the training dataset can improve $\Delta$LogLik even when the amount of data available is small. This is consistent with previous work such as \citet{hale-etal-2019-text}, showing that genre matters when it comes to cognitive modeling. On the other hand, observe that the 3-, 4-, and 5-gram models trained on LM1B outperform Transformer-XL and rival XLNet on $\Delta$LogLik for all three reading time metrics, despite having a much simpler model architecture. Given that LM1B (4 GB) is much bigger than WikiText-103 (515 MB), the training corpus for Transformer-XL, this observation shows that a large dataset can dramatically enhance the psycholinguistic modeling capabilities of an otherwise simple architecture. 
   
   It is worth noting that dataset size and quality are both important factors for $\Delta$LogLik. \autoref{fig:datasetsize_deltaloglik} shows the average $\Delta$LogLik for first fixation duration attained by models trained on each dataset, excluding XLM. There, we find that GPT-2 outperforms the other models in terms of psycholinguistic predictive power, including those that are more architecturally sophisticated, namely Transformer-XL and XLNet. Although WebText (40 GB) is not the largest dataset, as it is smaller than XLNet's training corpus (158 GB), it was constructed in a more curated approach than the other datasets. The wide variety of document types included in WebText likely makes it a higher-quality training corpus than XLNet's corpus or LM1B, allowing GPT-2 to surpass other models.
   
   \subsection{Factors Affecting PNC}
   
   Likewise, \autoref{fig:datasetsize_pnc} depicts the relationship between dataset size and PNC. Here, we find that the PNC consistently lies between 0.55 and 0.65 for the larger datasets, but is significantly lower for the smaller datasets. While the lower values for WikiText-2 and Penn Treebank show that smaller datasets generally produce models with lower PNC, the small variance in PNC for the larger datasets is suggestive of diminishing returns in PNC when training corpora are sufficiently large. 
   
   
   \begin{figure}
        \centering
        \includegraphics{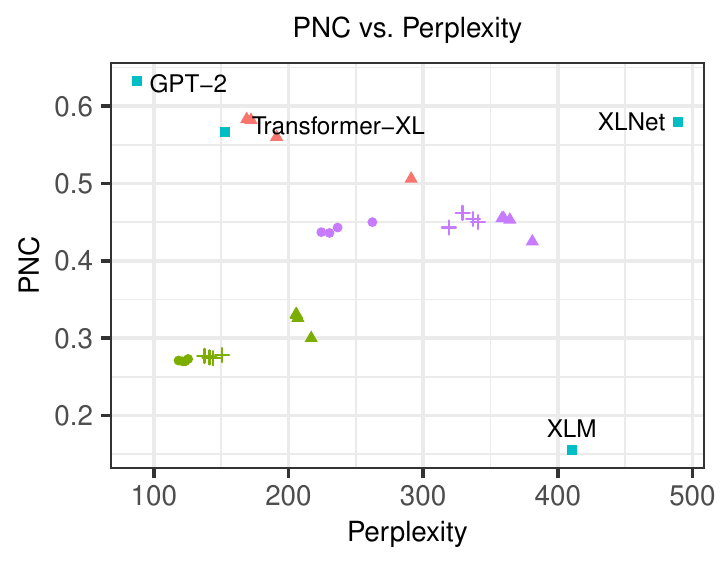}
        \caption{There is no correlation between perplexity and PNC ($\rho = \text{.222}$).}
        \label{fig:ppl_pnc}
    \end{figure}
    \begin{figure}[h]
        \centering
        \includegraphics{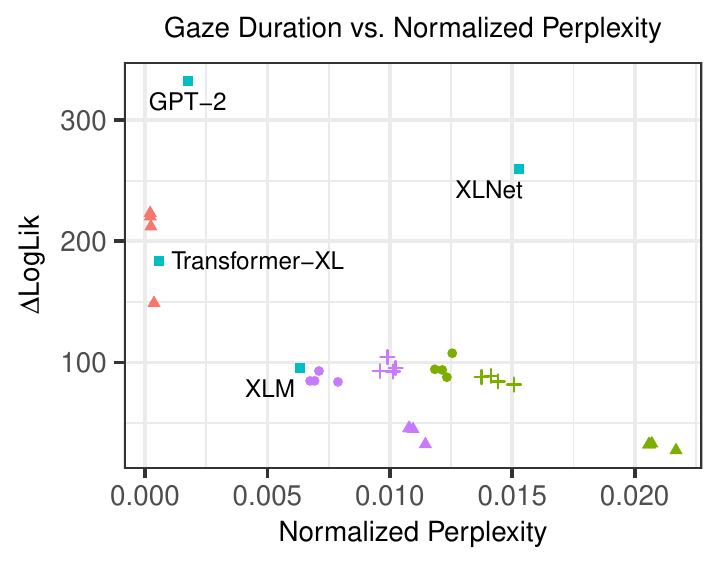}
        \caption{Unlike PNC, normalized perplexity does not exhibit a strong relation with $\Delta$LogLik.}
        \label{fig:normalized_ppl}
    \end{figure}
   
   Figures \ref{fig:ppl_pnc} and \ref{fig:normalized_ppl} compare PNC to perplexity-based metrics. \autoref{fig:ppl_pnc} shows that PNC cannot be predicted from perplexity, highlighting the distinctness of the two metrics. \autoref{fig:normalized_ppl} plots the $\Delta$LogLik of gaze duration against \textit{normalized perplexity} \citep{martiInfluenceVocabularySize2001}, defined as perplexity divided by vocabulary size. The relationship between normalized perplexity and $\Delta$LogLik appears to be slightly stronger than perplexity alone, but not as strong as the relationship between PNC and $\Delta$LogLik. This suggests that the benefits of using PNC over perplexity cannot be replicated simply by adjusting perplexity for differences in vocabulary size.
    
    \subsection{Evaluating Surprisal Estimates}

    While perplexity is the standard metric for evaluating surprisal estimates produced by language models, we have argued throughout this paper that perplexity is not a reliable metric of language modeling performance. In addition to the difficulty presented by perplexity for comparing models with different training conditions, especially pre-trained models, perplexity has been recently shown to be a poor predictor of a language model's ability to capture generalizations about natural language grammar \citep{ekLanguageModelingSyntactic2019,huSystematicAssessmentSyntactic2020}. Along those lines, in \autoref{fig:ppl_pnc} we have seen that perplexity is a poor predictor of PNC. Taken together, these observations indicate that perplexity does not capture the extent to which language models exhibit human-like behavior. Instead, alternative metrics like PNC or \citeauthor{huSystematicAssessmentSyntactic2020}'s (\citeyear{huSystematicAssessmentSyntactic2020}) syntactic generalization score explicitly assess the degree to which language models behave like humans, without sensitivity to training conditions or other artifacts of the model.
    
    Our arguments about the reliability of PNC as a measure of language modeling performance raise an interesting question about the role of language models in psycholinguistic modeling. Within the psycholinguistic modeling literature, language models are often viewed as statistical estimators of corpus-based frequencies (e.g., \citealp{smithClozeNoCigar2011}). While $n$-gram models can certainly be understood this way, the complex, increasingly opaque models used in neural NLP do not readily lend themselves to this interpretation. Instead, the literature on pre-training has recast language modeling as a generalization of NLP training objectives \citep{radfordLanguageModelsAre2019}, and pre-trained models have been shown to encode a wide range of linguistic information beyond word and $n$-gram frequencies (see \citealp{rogersPrimerBERTologyWhat2020} for an overview). Therefore, while \citet{frissonEffectsContextualPredictability2005} and \citet{smithClozeNoCigar2011} have argued that corpus-based frequencies are not as suitable for psycholinguistic modeling as Cloze probabilities, our results indicate that pre-trained models with a high PNC such as GPT-2 may capture a notion of predictability that more closely estimates subjective predictability than empirical probabilities, allowing them to serve as better psycholinguistic models.
    
    
    \section{Conclusion}
    \label{sec:conclusion}
    
    The results reported here suggest that for the purpose of modeling reading times, perplexity does not adequately reveal the relationship between language modeling performance and psycholinguistic modeling performance, especially when vocabulary cannot be controlled for. In contrast, PNC has proven to be a much better predictor of the psychometric capabilities of our language models. This finding is consistent with observations by \citet{frissonEffectsContextualPredictability2005} and \citet{smithClozeNoCigar2011} that Cloze probabilities predict self-paced reading times better than corpus probabilities do. In addition to allowing models with different vocabularies to be compared with one another, PNC is much more strongly correlated with psycholinguistic modeling performance than perplexity, as demonstrated by our results on $\Delta$LogLik. More generally, as there is little to no correlation between perplexity and PNC, PNC can serve as a good supplement to perplexity for language model evaluation, providing information about model behavior that is not captured by the latter. 
    
    We have also shown that model architecture, training dataset size, and training dataset composition all contribute substantially to the psycholinguistic modeling capabilities of language models. In particular, the importance of corpus size and composition is reflective of trends in transfer learning, in which advances in downstream NLP tasks are made by using language models to extract general linguistic information from large corpora. Our analysis has shown that large corpora have the potential to provide considerable amounts of linguistic knowledge even through simple model architectures, as in the case of the LM1B $n$-grams.
    
    As new pre-trained language models are developed, especially with custom vocabularies that make a direct comparison of perplexities impossible, metrics such as PNC can serve as a valuable tool for assessing the quality of language models. In establishing a strong relationship between PNC and $\Delta$LogLik, we have demonstrated that PNC scores convey a psychometrically relevant notion of language model quality that directly measures the degree to which language models exhibit human-like behavior. Reliable metrics like PNC, which are robust to variations in model setup, have the potential to greatly improve our ability to understand the relationship between language models and language.
    
    \section*{Acknowledgments}
    
    We would like to thank Adam Goodkind, Klinton Bicknell, and Vera Demberg for helping us reproduce \citeauthor{goodkindPredictivePowerWord2018}'s (\citeyear{goodkindPredictivePowerWord2018}) analysis; Alan Kennedy for providing us the Dundee Corpus data; Christopher Geissler for his helpful discussion and assistance in statistical modeling; and the reviewers for their feedback.

    \bibliographystyle{acl_natbib}
    \bibliography{emnlp2020}

\end{document}